\documentclass[10pt,twocolumn,letterpaper]{article}

\usepackage{cvpr}
\usepackage{times}
\usepackage{epsfig}
\usepackage{graphicx}
\usepackage{amsmath}
\usepackage{amssymb}
\usepackage{enumitem}
\usepackage{wrapfig}

\usepackage[pagebackref=true,breaklinks=true,letterpaper=true,colorlinks,bookmarks=false]{hyperref}

\cvprfinalcopy % *** Uncomment this line for the final submission

\def\cvprPaperID{5} % *** Enter the CVPR Paper ID here

\ifcvprfinal\fi
\begin{document}

%%%%%%%%% TITLE
\title{BlazeFace: Sub-millisecond Neural Face Detection on Mobile GPUs}

\author{Valentin Bazarevsky \quad Yury Kartynnik \quad Andrey Vakunov \quad Karthik Raveendran \quad Matthias Grundmann\\
Google Research\\
1600 Amphitheatre Pkwy, Mountain View, CA 94043, USA\\
{\tt\small \{valik, kartynnik, vakunov, krav, grundman\}@google.com}
}

\newcommand{\X}{$\times$}

% #1: resolution
% #2: in channel count
% #3: out channel count
% #4: stride
\newcommand{\singleblazerow}[4]{%
  Single BlazeBlock & #1\X#1\X#2
  & 5\X5\X#2\X1 #4 \\
  && 1\X1\X#2\X#3 \\ \hline
}

% #1: resolution
% #2: in channel count
% #3: stride
\newcommand{\doubleblazerow}[3]{%
  Double BlazeBlock & #1\X#1\X#2
  & 5\X5\X#2\X1 #3 \\
  && 1\X1\X#2\X24 \\
  && 5\X5\X24\X1 \\
  && 1\X1\X24\X96 \\ \hline
}

\maketitle
%\thispagestyle{empty}

%%%%%%%%% ABSTRACT
\begin{abstract}
   We present BlazeFace, a lightweight and well-perfor\-ming face detector tailored for mobile GPU inference. It runs at a speed of 200--1000+ FPS  on flagship devices. This super-realtime performance enables it to be applied to any augmented reality pipeline that requires an accurate facial region of interest as an input for task-specific models, such as 2D/3D facial keypoint or geometry estimation, facial features or expression classification, and face region segmentation. Our contributions include a lightweight feature extraction network inspired by, but distinct from MobileNetV1/V2, a GPU-friendly anchor scheme modified from Single Shot MultiBox Detector (SSD), and an improved tie resolution strategy alternative to non-maximum suppression.
\end{abstract}

\section{Introduction}

In recent years, a variety of architectural improvements in deep networks (\cite{SSD,FastRCNN,FasterRCNN}) have enabled real-time object detection. In mobile applications, this is usually the first step in a video processing pipeline, and is followed by task-specific components such as segmentation, tracking, or geometry inference. Therefore, it is imperative that the object detection model inference runs as fast as possible, preferably with the performance much higher than just the standard real-time benchmark.

We propose a new face detection framework called BlazeFace that is optimized for inference on mobile GPUs, adapted from the Single Shot Multibox Detector (SSD) framework~\cite{SSD}. Our main contributions are:
\vspace{-\topsep}
\begin{enumerate}[leftmargin=*]
\setlength{\parskip}{0pt}
\setlength{\itemsep}{0pt plus 3pt}
  \item Related to the inference speed:
  \begin{enumerate}[leftmargin=*]
      \vspace{-\topsep}
      \setlength{\parskip}{0pt}
      \setlength{\itemsep}{0pt plus 2pt}
      \item[1.1.] A very compact feature extractor convolutional neural network related in structure to MobileNetV1/V2~\cite{MobileNetV1, MobileNetV2}, designed specifically for lightweight object detection.
      \item[1.2.] A novel GPU-friendly anchor scheme modified from SSD~\cite{SSD}, aimed at effective GPU utilization. \emph{Anchors}~\cite{FasterRCNN}, or \emph{priors} in SSD terminology, are predefined static bounding boxes that serve as the basis for the adjustment by network predictions and determine the prediction granularity.
    \end{enumerate}
  \item Related to the prediction quality: A tie resolution strategy alternative to non-maximum suppression~\cite{SSD,FastRCNN,FasterRCNN} that achieves stabler, smoother tie resolution between overlapping predictions.
\end{enumerate}
\vspace{-\topsep}

\section{Face detection for AR pipelines} \label{face}

While the proposed framework is applicable to a variety of object detection tasks, in this paper we focus on detecting faces in a mobile phone camera viewfinder. We build separate models for the front-facing and rear-facing cameras owing to the different focal lengths and typical captured object sizes. 

In addition to predicting axis-aligned face rectangles, our BlazeFace model produces 6 facial keypoint coordinates (for eye centers, ear tragions, mouth center, and nose tip) that allow us to estimate face rotation (roll angle). This enables passing a rotated face rectangle to later task-specific stages of the video processing pipeline, alleviating the requirement of significant translation and rotation invariance in subsequent processing steps (see Section \ref{sect:applications}).

\section{Model architecture and design}

BlazeFace model architecture is built around four important design considerations discussed below.

\paragraph{Enlarging the receptive field sizes.}

While most of the modern convolutional neural network architectures (including both MobileNet \cite{MobileNetV1,MobileNetV2} versions) tend to favor 3\X3 convolution kernels everywhere along the model graph, we note that the depthwise separable convolution computations are dominated by their pointwise parts. On an $s\times s \times c$ input tensor, a $k \times k$ depthwise convolution involves $s^2 c k^2$ multiply-add operations, while the subsequent $1 \times 1$ convolution into $d$ output channels is comprised of $s^2 c d$ such operations, within a factor of $d / k^2$ of the depthwise part.

In practice, for instance, on an Apple iPhone X with the Metal Performance Shaders implementation~\cite{MetalShaders}, a 3\X3 depthwise convolution in 16-bit floating point arithmetic takes 0.07 ms for a 56\X56\X128 tensor, while the subsequent 1\X1 convolution from 128 to 128 channels is 4.3$\times$ slower at 0.3 ms (this is not as significant as the pure arithmetic operation count difference due to fixed costs and memory access factors).

\begin{figure}
  \includegraphics[width=\linewidth]{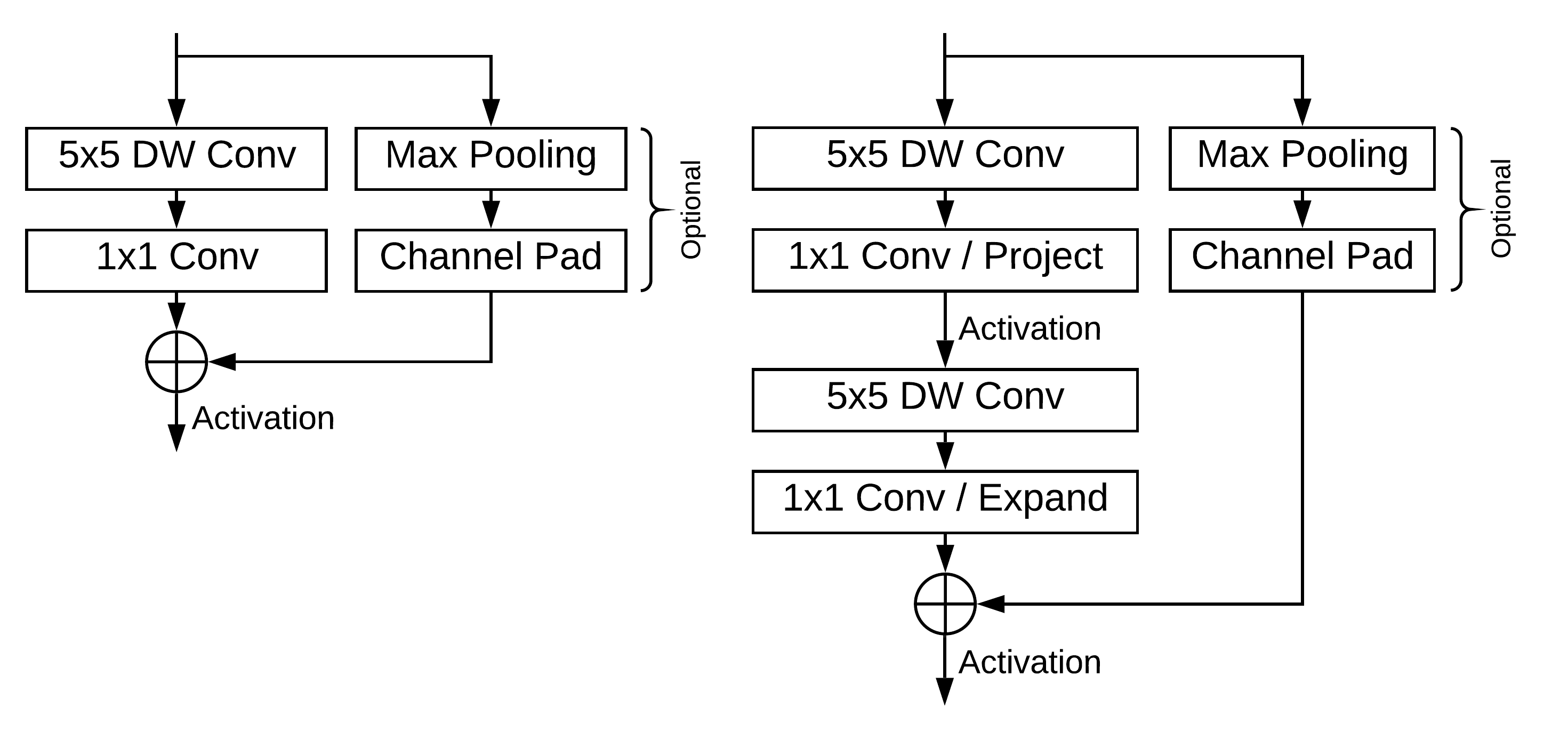}
  \caption{BlazeBlock (left) and double BlazeBlock}
  \label{fig:bb_blaze_block}
\end{figure}

This observation implies that increasing the kernel size of the depthwise part is relatively cheap. We employ 5\X5 kernels in our model architecture bottlenecks, trading the kernel size increase for the decrease in the total amount of such bottlenecks required to reach a particular receptive field size (Figure~\ref{fig:bb_blaze_block}).

A MobileNetV2~\cite{MobileNetV2} bottleneck contains subsequent depth-increasing \emph{expansion} and depth-decreasing \emph{projection} pointwise convolutions separated by a non-linearity. To accommodate for the fewer number of channels in the intermediate tensors, we swap these stages so that the residual connections in our bottlenecks operate in the ``expanded'' (increased) channel resolution.

Finally, the low overhead of a depthwise convolution allows us to introduce another such layer between these two pointwise convolutions, accelerating the receptive field size progression even further. This forms the essence of a \emph{double BlazeBlock} that is used as the bottleneck of choice for the higher abstraction level layers of BlazeFace (see Figure~\ref{fig:bb_blaze_block}, right).

\paragraph{Feature extractor.}
For a specific example, we focus on the feature extractor for the front-facing camera model. It has to account for a smaller range of object scales and therefore has lower computational demands.
The extractor takes an RGB input of 128\X128 pixels and consists of a 2D convolution followed by 5 single BlazeBlocks and 6 double BlazeBlocks (see Table~\ref{tbl:architecture} in Appendix A for the full layout). The highest tensor depth (channel resolution) is 96, while the lowest spatial resolution is 8\X8 (in contrast to SSD, which reduces the resolution all the way down to 1\X1).

\paragraph{Anchor scheme.}
\begin{figure}
  \includegraphics[width=\linewidth]{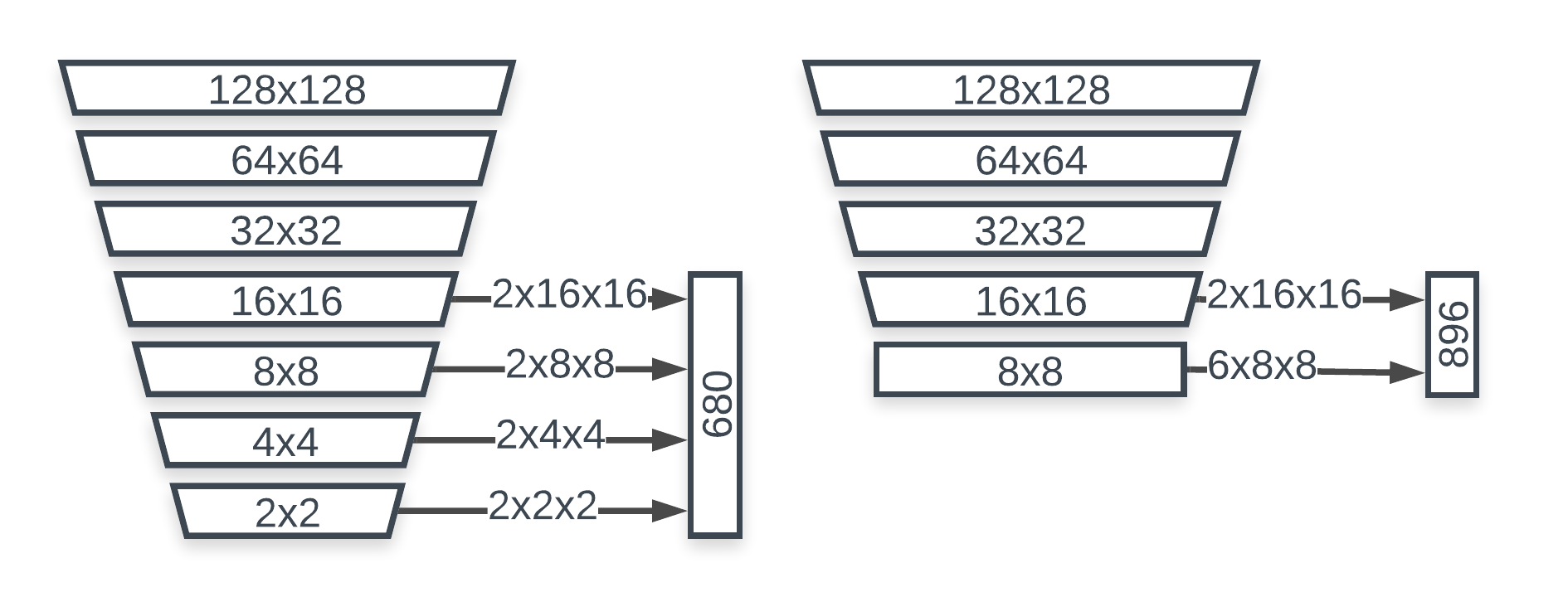}
  \caption{Anchor computation: SSD (left) vs. BlazeFace}
  \label{fig:anchors}
\end{figure}

SSD-like object detection models rely on pre-defined fixed-size base bounding boxes called \emph{priors}, or \emph{anchors} in Faster-R-CNN \cite{FasterRCNN} terminology. A set of regression (and possibly classification) parameters such as center offset and dimension adjustments is predicted for each anchor. They are used to adjust the pre-defined anchor position into a tight bounding rectangle.

It is a common practice to define anchors at multiple resolution levels in accordance with the object scale ranges. Aggressive downsampling is also a means for computational resource optimization. A typical SSD model uses predictions from 1\X1, 2\X2, 4\X4, 8\X8, and 16\X16 feature map sizes. However, the success of the Pooling Pyramid Network (PPN) architecture \cite{PPN} implies that additional computations could be redundant after reaching a certain feature map resolution.

A key feature specific to GPU as opposed to CPU computation is a noticeable fixed cost of dispatching a particular layer computation, which becomes relatively significant for deep low-resolution layers inherent to popular CPU-tailored architectures. As an example, in one experiment we observed that out of 4.9 ms of MobileNetV1 inference time only 3.9 ms were spent in actual GPU shader computation.

Taking this into consideration, we have adopted an alternative anchor scheme that stops at the 8\X8 feature map dimensions without further downsampling (Figure \ref{fig:anchors}). We have replaced 2 anchors per pixel in each of the 8\X8, 4\X4 and 2\X2 resolutions by 6 anchors at 8\X8. Due to the limited variance in human face aspect ratios, limiting the anchors to the 1:1 aspect ratio was found sufficient for accurate face detection.

\paragraph{Post-processing.}
As our feature extractor is not reducing the resolution below 8\X8, the number of anchors overlapping a given object significantly increases with the object size. In a typical non-maximum suppression scenario, only one of the anchors ``wins'' and is used as the final algorithm outcome. When such a model is applied to subsequent video frames, the predictions tend to fluctuate between different anchors and exhibit temporal jitter (human-perceptible noise).

To minimize this phenomenon, we replace the \emph{suppression} algorithm with a \emph{blending} strategy that estimates the regression parameters of a bounding box as a weighted mean between the overlapping predictions. It incurs virtually no additional cost to the original NMS algorithm. 
For our face detection task, this adjustment resulted in a 10\% increase in accuracy.

We quantify the amount of jitter by passing several slightly offset versions of the same input image into the network and observing how the model outcomes (adjusted to account for the translation) are affected. After the described tie resolution strategy modification, the jitter metric, defined as the root mean squared difference between the predictions for the original and displaced inputs, dropped down by 40\% on our frontal camera dataset and by 30\% on a rear-facing camera dataset containing smaller faces.

\section{Experiments}

We trained our model on a dataset of 66K images. For evaluation, we used a private geographically diverse dataset consisting of 2K images. For the frontal camera model, only faces that occupy more than 20\% of the image area were considered due to the intended use case (the threshold for the rear-facing camera model was 5\%).

 The regression parameter errors were normalized by the inter-ocular distance (IOD) for scale invariance, and the median absolute error was measured to be 7.4\% of IOD. The jitter metric evaluated by the procedure mentioned above was 3\% of IOD.

Table \ref{tbl:benchmarks} shows the average precision (AP) metric~\cite{PASCAL} (with a standard 0.5 intersection-over-union bounding box match threshold) and the mobile GPU inference time for the proposed frontal face detection network and compares it to a MobileNetV2-based object detector with the same anchor coding scheme (MobileNetV2-SSD). We have used TensorFlow Lite GPU \cite{TFLiteGPU} in 16-bit floating point mode as the framework for inference time evaluation.

\begin{table}[ht]
\centering
\begin{tabular}{|l|c|c|c|c|}
\hline
Model & Average & Inference Time, ms \\
 & Precision & (iPhone XS) \\ \hline
MobileNetV2-SSD & 97.95\% & 2.1 \\ \hline
Ours & \textbf{98.61}\% & \textbf{0.6} \\ \hline
\end{tabular}
\vskip 1ex
\caption{Frontal camera face detection performance}
\label{tbl:benchmarks}
\end{table}

Table \ref{tbl:more_speed} gives a perspective on the GPU inference speed for the two network models across more flagship devices.

\begin{table}[ht]
\centering
\begin{tabular}{|l|c|c|}
\hline
Device & MobileNetV2-SSD, ms & Ours, ms \\ \hline
Apple iPhone 7 & 4.2 & \textbf{1.8} \\ \hline
Apple iPhone XS & 2.1 & \textbf{0.6} \\ \hline
Google Pixel 3 & 7.2 & \textbf{3.4} \\ \hline
Huawei P20 & 21.3 & \textbf{5.8} \\ \hline
Samsung Galaxy S9+ & 7.2 & \textbf{3.7} \\
(SM-G965U1) & & \\ \hline
%Samsung S10 & 16.3 & \textbf{12.1} \\ \hline  % Replaced with S9 as Jet has not been optimized for this model
\end{tabular}
\vskip 1ex
\caption{Inference speed across several mobile devices}
\label{tbl:more_speed}
\end{table}

Table \ref{tbl:regression} shows the amount of degradation in the regression parameter prediction quality that is caused by the smaller model size. As explored in the following section, this does not necessarily incur a proportional degradation of the whole AR pipeline quality.

\begin{table}[ht]
\centering
\begin{tabular}{|l|c|c|c|c|}
\hline
Model &  Regression & Jitter \\
 & error & metric \\ \hline
MobileNetV2-SSD & \textbf{7.4}\% & \textbf{3.6}\% \\ \hline
Ours & 10.4\% & 5.3\% \\ \hline
\end{tabular}
\vskip 1ex
\caption{Regression parameters prediction quality}
\label{tbl:regression}
\end{table}

\section{Applications} \label{sect:applications}

The proposed model, operating on the full image or a video frame, can serve as the first step of virtually any face-related computer vision application, such as 2D/3D facial keypoints, contour, or surface geometry estimation, facial features or expression classification, and face region segmentation. The subsequent task in the computer vision pipeline can thus be defined in terms of a proper facial crop. Combined with the few facial keypoint estimates provided by BlazeFace, this crop can be also rotated so that the face inside is centered, scale-normalized and has a roll angle close to zero. This removes the requirement of significant translation and rotation invariance from the task-specific model, allowing for better computational resource allocation.

\begin{wrapfigure}{r}{0pt}
  \centering
  \includegraphics[width=0.49\linewidth]{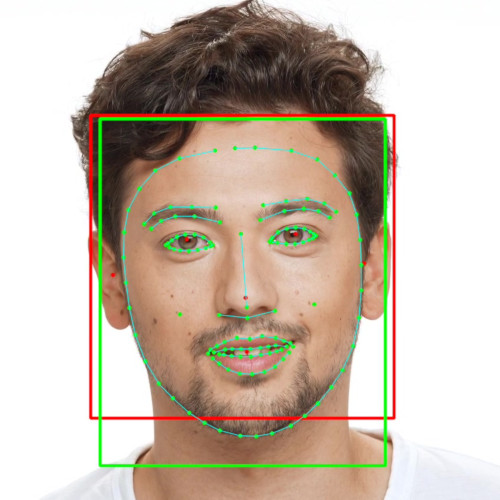}
  \caption{\protect\centering Pipeline example (best viewed in color).\protect\linebreak Red: BlazeFace output. Green: Task-specific model output.}
  \label{fig:contours}
\end{wrapfigure}

We illustrate this pipelined approach with a specific example of face contour estimation. In Figure \ref{fig:contours}, we show how the output of BlazeFace, i.e. the predicted bounding box and the 6 keypoints of the face (red), are further refined by a more complex face contour estimation model applied to a slightly expanded crop. The detailed keypoints yield a finer bounding box estimate (green) that can be reused for tracking in the subsequent frame without running the face detector. To detect failures of this computation saving strategy, the contours model can also detect whether the face is indeed present and reasonably aligned in the provided rectangular crop. Whenever that condition is violated, the BlazeFace face detector is run on the whole video frame again.

The technology described in this paper is driving major AR self-expression applications and AR developer APIs on mobile phones.

{\small
\bibliographystyle{ieee_fullname}
\bibliography{blazebib}
}

\newpage
\section*{Appendix A. Feature extraction network architecture}

\begin{table}[h!]
    \centering
    \begin{tabular}{|l|l|l|}
    \hline
        Layer/block & Input size & Conv. kernel sizes \\ \hline
        Convolution & 128\X128\X3 & 5\X5\X3\X24 (stride 2)\\ \hline
        \singleblazerow{64}{24}{24}{}
        \singleblazerow{64}{24}{24}{}
        \singleblazerow{64}{24}{48}{(stride 2)}
        \singleblazerow{32}{48}{48}{}
        \singleblazerow{32}{48}{48}{}
        \doubleblazerow{32}{48}{(stride 2)}
        \doubleblazerow{16}{96}{}
        \doubleblazerow{16}{96}{}
        \doubleblazerow{16}{96}{(stride 2)}
        \doubleblazerow{8}{96}{}
        \doubleblazerow{8}{96}{}
    \end{tabular}
    \vskip 1ex
    \caption{BlazeFace feature extraction network architecture}
    \label{tbl:architecture}
\end{table}

\end{document}